\documentclass[manuscript,screen]{acmart}
\usepackage[ruled,vlined,linesnumbered]{algorithm2e}
\usepackage{algpseudocode}
\usepackage[]{algorithm2e}
\AtBeginDocument{%
  \providecommand\BibTeX{{%
    \normalfont B\kern-0.5em{\scshape i\kern-0.25em b}\kern-0.8em\TeX}}}

\setcopyright{acmcopyright}
\copyrightyear{2022}
\acmYear{2022}
\acmDOI{}

\acmJournal{TEAC}

\begin{document}

\title{Estimating defection in subscription‐type markets: empirical analysis from the scholarly publishing industry}

\author{Michael Roberts}
\author{J. Ignacio Deza}
\authornote{Main contributors and corresponding authors. }
\email{ignacio.deza@uwe.ac.uk}
\orcid{0000-0001-9551-9718}
\author{Hisham Ihshaish}
\authornotemark[1]
\email{hisham.ihshaish@uwe.ac.uk}
\orcid{0000-0001-5530-4894}
\author{Yanhui Zhu}
\affiliation{%
  \institution{\newline School of Computing and Creative Technologies and Bristol Business School, UWE Bristol.}
  \streetaddress{Frenchay Campus, Coldharbour Ln}
  \city{Bristol}
  \country{UK}
  \postcode{BS16 1QY}
}

\renewcommand{\shortauthors}{Roberts, M. et al.}

\begin{abstract}
We present the first empirical study on customer churn prediction in the scholarly publishing industry. The study examines our proposed method for prediction on a customer subscription data over a period of 6.5 years, which was provided by a major academic publisher. We explore the subscription-type market within the context of customer defection and modelling, and provide analysis of the business model of such markets, and how these characterise the academic publishing business. The proposed method for prediction attempts to provide inference of customer's likelihood of defection on the basis of their re-sampled use of provider resources --in this context, the volume and frequency of content downloads. We show that this approach can be both accurate as well as uniquely useful in the business-to-business context, with which the scholarly publishing business model shares similarities. The main findings of this work suggest that whilst all predictive models examined, especially ensemble methods of machine learning, achieve substantially accurate prediction of churn, nearly a year ahead, this can be furthermore achieved even when the specific behavioural attributes that can be associated to each customer probability to churn are overlooked. Allowing as such highly accurate inference of churn from minimal possible data. We show that modelling churn on the basis of re-sampling customers' use of resources over subscription time is a better (simplified) approach than when considering the high granularity that can often characterise consumption behaviour.  

\end{abstract}





\maketitle

\section{Introduction}
\label{sec:1}

It is long-established that retaining existing customer-base is critical to maintaining business profitability. The literature in fact is replete with studies on the effects of the so-called customer `churn' --otherwise referred to as attrition or defection-- on revenue, and subsequently on modelling it. Over hundreds of such studies only in the last decade, the consensus is that customer defection is distinctly problematic as the cost to acquire new customers --on average-- is greater than that of \textit{serving} existing ones. 

This, especially challenging in saturated markets, has been reported consistently since it was first analysed empirically in \cite{Reichheld1990ZeroDQ} --organisations must prioritise ``zero defections'', as for all that the magnitude of revenue may vary across the different industries, the pattern nonetheless holds: profits rise strikingly as churn rate is (even when marginally) reduced. Added to the salient disparity between the cost of attracting new customers relative to that of retaining existing ones -- which in some studies has been estimated to be five to six-fold greater \cite{Verbeke} --, long-term customers tend to be generally more profitable: they develop some sort of brand loyalty and consequently, on the one hand are less likely to defect (being less susceptible to competitive marketing), and can often serve as a marketing broker through their positive word-of-mouth referrals, on another.

In general, the catch-all term of customer churn -- which, in its broadest sense can be defined as the outflow of customers from a business customer base \cite{Baesensbook} -- is agreed to have a significant financial impact on organisations. The case of subscription-type markets --where customers allocate their service to one provider for a given time period-- is no different. Dawes (2004) emphasises that contrary to ``repertoire'' markets, where customers buy from a repertoire of multiple providers \cite{sharp99}, defection in subscription-type markets can be even more costly, as customer acquisition in such a market typically tends to include additional up-front costs such as credit checks, contractual and data processing, etc. \cite{Dawes}.   

Whatever is the market-type, the amount of revenue that can be lost from customer defection must be huge. This probably explains the (relatively) ample literature that can be found on customer retention and customer churn modelling, and prediction. The first largely looked at the determinants of customer defection, e.g. \cite{ESHGHI200793, SEO2008182,VERBEKE20112354} whereas the latter, being the subject of interest in this paper, is concerned with predicting customers' likelihood to defect as an integral part of any business retention strategy. Such strategy predominantly depends on an effective prediction modelling of churn. That is, predictive models --methods of machine learning and statistical inference-- are designed to identify customers with high probability of defection in the future. This in turn is consequently used to shape an effective retention marketing and investment strategy, e.g. customer-centric retention marketing campaigns (their impact is reported in \cite{Measuringimpact}) and price discrimination (otherwise known as dynamic pricing) which have shown to be largely effective in reducing defection rates \cite{CAPPONI2021102069, ESTEVES201439}. 

On this there's clearly no shortage of material, with varying degrees of reported accuracy, we found hundreds of published work only in the last decade. A simple search query on \textit{Google Scholar} for the expression `customer churn prediction' returned approximately twenty six thousand items at the time of writing this manuscript\footnote{Result can be retrieved here: https://bit.ly/3DrZPaA} --this paper is another addition to this vast portfolio of published work on churn prediction, albeit presenting a novel modelling approach and promising results to the problem. Examples include applications to: (1) the telecommunication market as in \cite{Qureshi2013TelecommunicationSC, Kim2014ImprovedCP, Ahmad2019CustomerCP, Li2020CustomerCP}, (2) the banking industry \cite{bilal2016predicting, keramati2016developing, dias2020machine}, (3) the insurance industry \cite{morik2004analysing, gunther2014modelling, scriney2020predicting}, (4) subscription-type markets \cite{richter2010predicting, hudaib2015hybrid, kolomiiets2021customer}, as well as to a wide range of markets such as software-as-a-service \cite{ge2017customer}, internet broadband service \cite{huang2009customer}, the video-on-demand industry \cite{huang2009customer} and online gaming \cite{onlinegaming}. 

Compared to the mass of studies addressing churn prediction in the business-to-customer (B2C) markets, the business-to-business (B2B) context has received much less attention from both scholars and practitioners. The scholarly publishing market is one example; largely a B2B market, there is no published research, to the best of our knowledge, addressing the prediction of customer defection and retention. 

\newpage

The following section provides an overview of the market, its size and characteristics, whereas the reminder of this paper will focus on setting forth methods to estimating customer churn in the industry, and is organised as follows: Section \ref{overview} provides an overview of the scholarly publishing industry, their business model as well as the effect of customer churn in the business. In Section \ref{data} we describe the data we analyse and define consumption measures typically used in the industry; in Section \ref{methods} we discuss the methodology for the presented study and the approach to choices for prediction modelling. Section \ref{results} provides a summary of results and related discussion and finally conclusions are summarised in Section \ref{conclusion}.

\section{The Scholarly Publishing Industry: An overview}
\label{overview}
The centrality of predicting customer defection to B2B markets' customer retention efforts should be of no less substance. In fact it is largely accepted that it can be even more indispensable \cite{gordini2017customers}, especially so as such businesses typically rely on a fewer number of customers on the one hand, and these are usually more valuable --purchase volume and frequency-- on another \cite{rauyruen2007relationship}. Loosing one customer may result damaging, and conversely could be the financial reward as a result of their retention. For more discussion on this, reader is advised to refer to \cite{figalist2019customer}. 

The scholarly (or scientific and technical) publishing industry is one example of such B2B businesses providing services to organisational customers, e.g. libraries, universities, colleges and schools ---big and small, but also governmental and non-governmental organisations---. This is due to the fact that readers of scientific literature are mostly affiliated to an institution--academic or otherwise--, just like the authors of this manuscript, and those who are not affiliated are unlikely to pay (typically) high costs per article as offered by the publishers. It is in fact a common practice for such users to resort to requesting access to articles from their corresponding authors directly, or to try and gain access through research social networks, such as \textit{Research Gate}\footnote{https://researchgate.com},  where a version of the required material can be accessed free of charge. Although some material can be provided in print, most of the consumption of the scientific and technical contents takes place electronically, via a portal or by using an ``institutional access'' credentials to a wide range of online services. Authorised users are able to navigate almost transparently while the publisher is also able to keep record of the activities of each institution. Typically users pay an annual fee to gain access to publishers' provided contents, and they may let their subscription lapse if they decide to.

As with the B2B context in general, the scholarly publishing industry has received very little attention concerning customer churn analysis. This is despite the fact that its global market's size has been estimated at \$10.5 billion in 2020, with indicators for further growth in the following years, according to the ``Global Scientific \& Technical Publishing 2021-2025'' report published by \textit{Research and Markets}\footnote{Global Scientific \& Technical Publishing 2021-2025, Research and Markets. Access URL: shorturl.at\/akquW}. In their review, analysing additionally leading global scholarly publishers, such as Elsevier, IHS Markit, Springer Nature, Clarivate Analytics, John Wiley \& Sons, American Chemical Society, they highlight reports on ``university libraries canceling their journal subscription packages in 2020 and 2021, but most are still subscribing to individual journals based on usage/importance to the researchers and faculty''. That is why it is of a great urgency for such businesses to be able to understand, control and predict the use of their services, and accordingly design appropriate retention policies and campaigns.

 Typically, marketing departments of the publishing companies use a metric called ``cost-per-download'' (CPD) --similar to the cost-per-click (CPC) metric popular in the online advertising practice \cite{hu2016incentive,najafi2014cost}--, for their pricing policy, which is defined as the cost of the annual subscription fee divided by the number of content items downloaded by users at the customer institution. If CPD is too high, customers tend to consider reviewing their expenditure on the resource and ultimately, this may lead to a change in subscription to a lower price tier, or to the cancellation of the subscription altogether. This measure--although useful as a risk management tool, easy to calculate and to explain--may fall short of substance from a predictive point of view, precisely because it can lead to false positives (institutions that don't download much material but however are content with the subscription) or false negatives (institutions that have a high rate of download but are--nonetheless--thinking about churning), as a wide range of causes for variability (embedded in the CPD time series) are largely not considered.

In the following sections we explain the proposed approach to the analysis and prediction of customer churn in the industry, alongside data,  methods and results. The main question considered here is, by considering only row data provided in the form of customer time series of downloads (volume and temporal frequency), whether predictive modelling of churn can be possible, accurate and feasible. That is to say that the specific behavioural attributes that can be associated to each customer's (and their customers) probability to churn are to be overlooked. This can be uniquely useful in the B2B context due to the fact that such businesses are oftentimes required to handle their customers' data --of which their affiliated stakeholders' consumption behaviour can be inferred-- carefully, complying with the particular privacy and regulatory requirements. In fact the limited access to consumers' data, mostly a result of privacy and competitiveness concerns, is argued to have contributed to the relative shortage of research on customer churn prediction in the context of B2B, compared to that in B2C \cite{figalist2019customer}. The presented work therefore aims at examining predictive modelling of customer defection in the scholarly publishing business, whilst inferring from minimum possible data.

\section{Data}
\label{data}

Content is made available to subscribers both online on the publisher's web page as well as in the form of files which can be read offline (i.e. pdf or epub formats). To keep count of consumption, a metric called `Full Text Downloads'(FTDs) is typically used by publishers. An FTD is recorded each time a user accesses a web page with the content, or saves a copy of the offline version of the content for later reading. The data set used in this work is a compiled version of this recording. It contains daily data for 10279 users, the user names and countries were anonymised and appear as numbers. Data is additionally normalised so that the real number of downloads per user cannot be determined. The total time period covered in the data set is 6.5 years, although the starting and finishing dates for each client are not necessarily uniform, as these depend on their particular dates of subscription and cancellation, and can span from several months to several years. Access to data was facilitated under a non-disclosure agreement (NDA) requiring the anonymisation of the data, and the non disclosure of the identity of the Publisher or its subscribers. All data samples shown in this paper have been anonymised.

\begin{figure}
	\centering
		\includegraphics[width=0.65\linewidth]{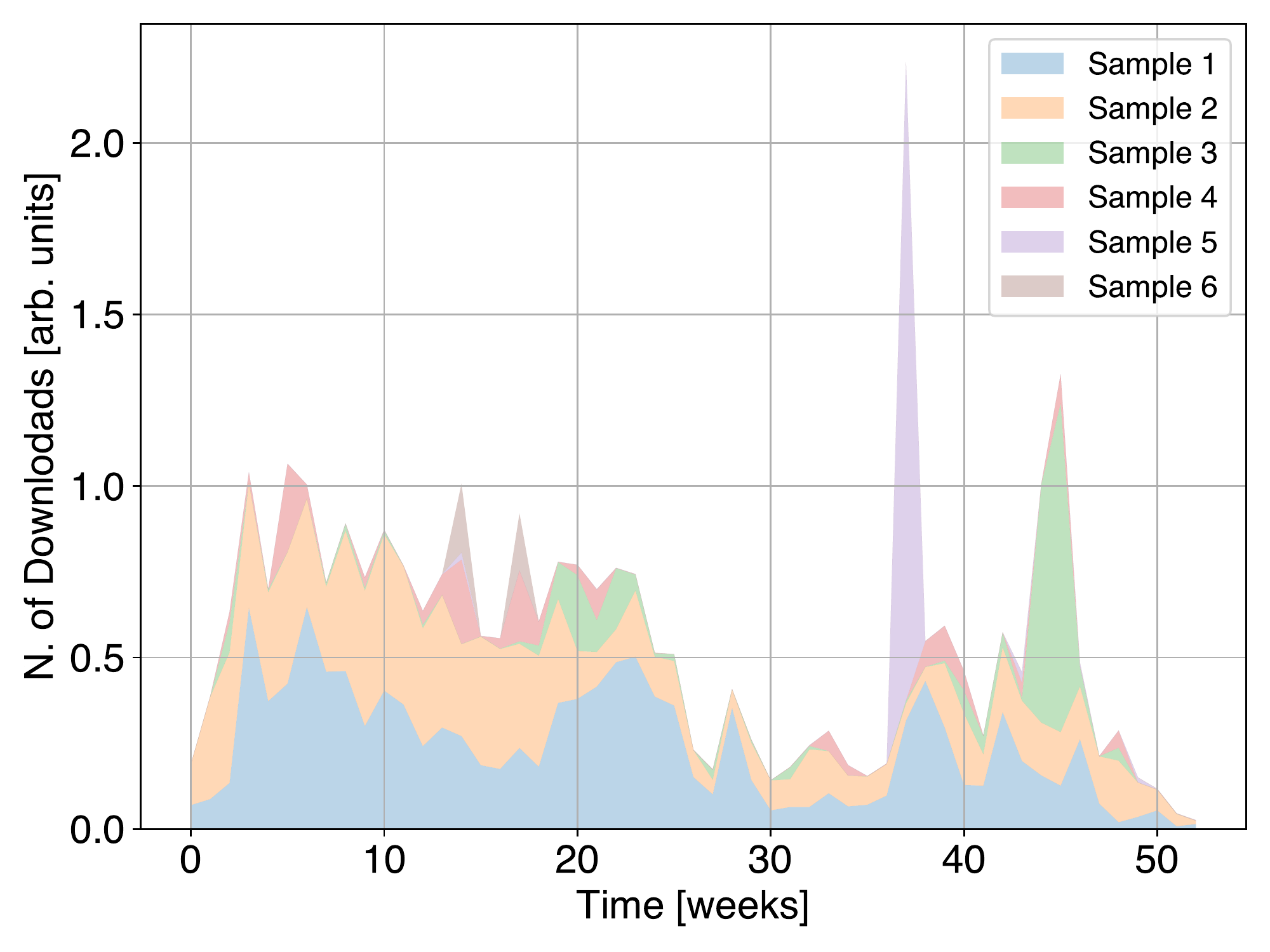}
	\caption{Daily stacked downloads for six randomly sampled customers for a randomly selected year in the data set, illustrating the range of use patterns seen in the data set. Data has been anonymised by normalising it to obscure the total number of downloads, where we additionally added a very small random number to each value.}
	\label{FIG:1}
\end{figure}

Figure \ref{FIG:1} shows the daily downloads for a random month for a sample customers. Some customers can have total daily downloads several orders of magnitude over others as users (customers) can be big or small institutions and yet only count as one consumer. Of course this is to say that there certainly is  complex download dynamics embedded in the time series of counts per customer. These dynamics may depend on a large amount of unknown variables, including the academic calendar for each institution, holidays, exams, etc\ldots However, we argue that none of these many variables are responsible for customer churn. Motivated by this, we will treat these variables as intrinsic variability in the time series.

%Further analysis has been performed on the data set. The time series have been de-seasonalised. As there are customers from all over the world, they don't follow the same academic years. This calendar is usually inverted in the southern hemisphere, and holidays vary in a high degree between countries or even between institutions. Because of this reason, from each time series, the There are strong seasonal trends in the data, with peaks in October and March, coinciding with the Western academic year. This trend was normalised by dividing usage numbers by the overall average value for day of year.

\begin{figure}[ht!]
	\centering
		\includegraphics[width=0.49\linewidth]{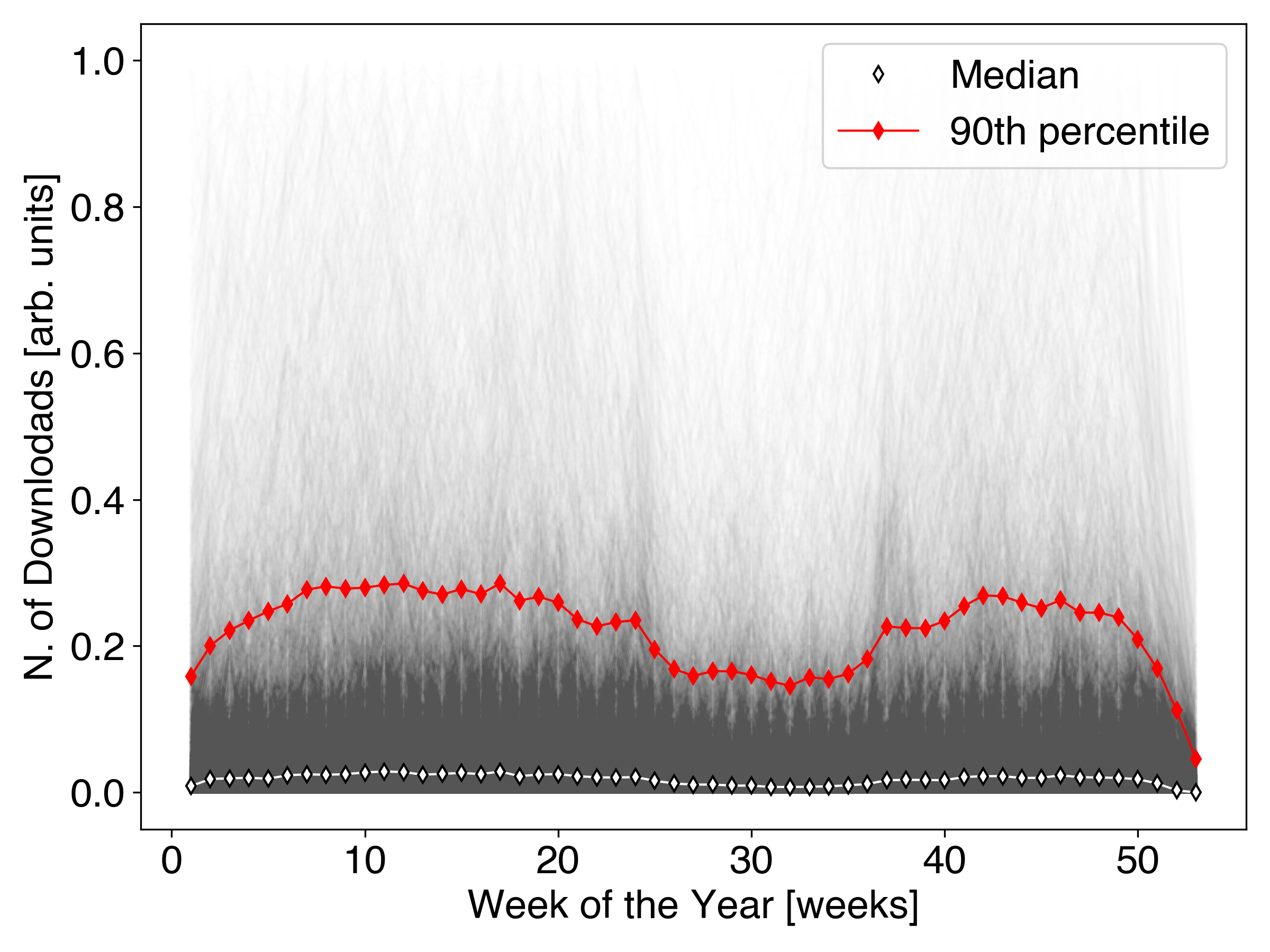}
		\includegraphics[width=0.49\linewidth]{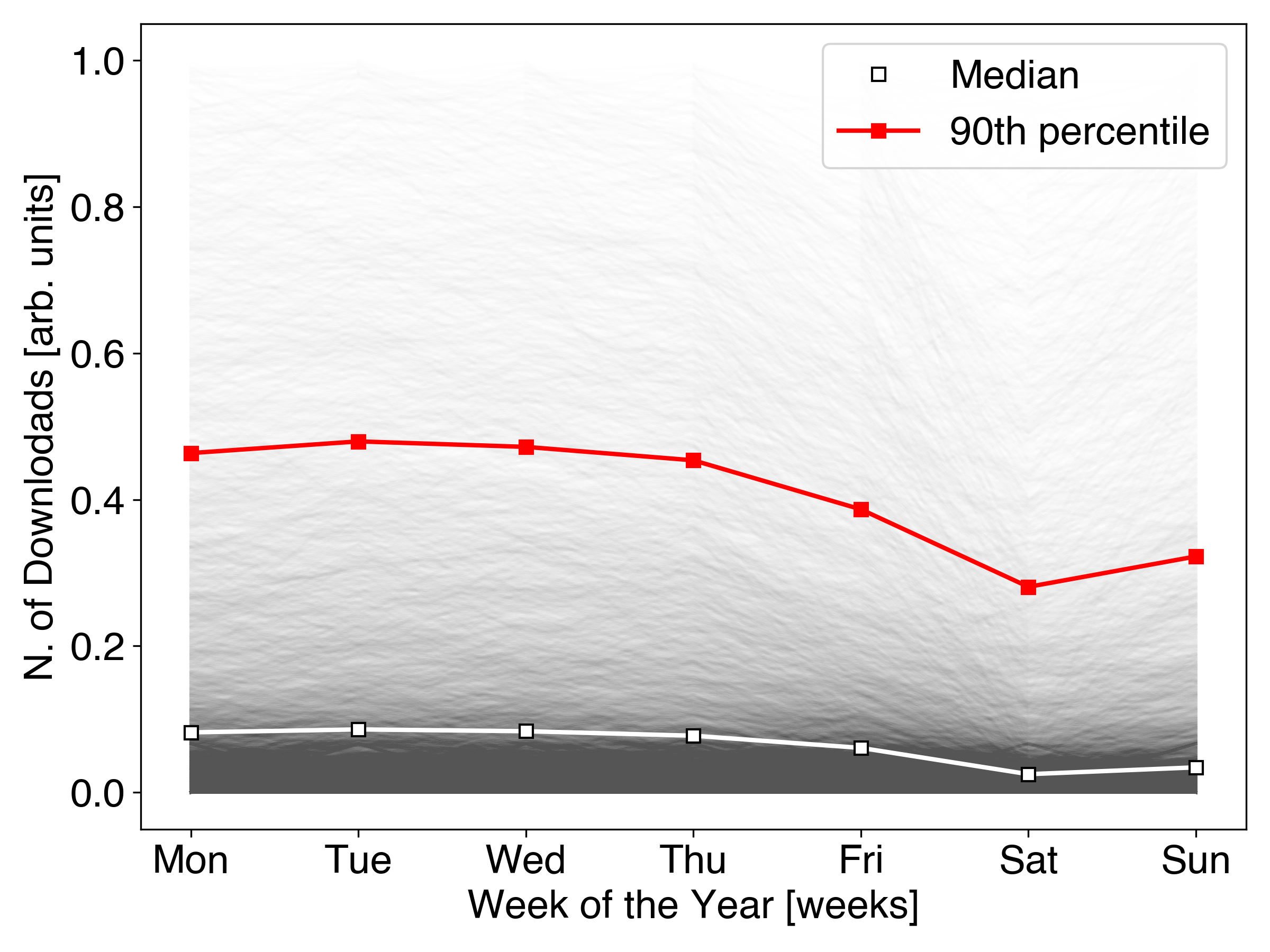}
	\caption{(left) Overall average weekly downloads, showing typical seasonality of customer behaviour. In gray the anonymised data, showing string weekly variations and very large excursions from the median (in white). Some values are much larger than the 90\% percentile (in red) which indicates that 90\% of the data is below the red line. (right) The weekly variability of the download patterns shows that ---on average--- highest volume of downloads happens on Wednesday whereas Saturday is the least active --where about half of the material is downloaded. It is interesting to note that this pattern appears to happen with both the high volume downloaders and the low volume ones.}
	\label{FIG:extraintraweek}
\end{figure}

In Figure \ref{FIG:extraintraweek} (left) we show an annual average of the FTDs in data. As the differences in downloading patterns and volume between institutions exhibit a notable variation, we used the median and percentiles instead of averages. In white one can see the that at least half of the customers do not download large volumes of content daily (see median). Furthermore, the 90th percentile shows that about 90\% of the customers exhibit downloading patterns of about 20\% of the highest downloaders. Average seasonal trends can additionally be seen. The minimum in download volume ---apart from the three last weeks of the year corresponding to the Holiday Season---is around week 30 of the year, which usually falls in late July or early August where downloads are low, as these  usually coincide with holidays in academic institutions across the Northern Hemisphere. The weekly seasonality was furthermore investigated and a clear trend emerged: downloads from Monday to Thursday are very similar and consistent, with a dip on Fridays; less activity on Saturdays and slightly higher on Sundays when arguably students and academics prepare for the next workweek.

\section{Methods}
\label{methods}

The first step is to classify the data into ``Active'' and ``Inactive'' customers. Because of user anonymity we have no access to the signing in or out dates for customers, and therefore we inferred these from the data itself. For this, we defined inactive customers as those who had not made any downloads for at least one month prior to the end of the covered date range. Of those, about 590 customers had never downloaded any content, arguably because they cancelled their subscription immediately and/or never used the service. We decided to exclude these samples from the study. The rest of the data set was labeled into two groups of approximately the same size (5141 active and 4552 inactive accounts). The inactive accounts were defined as those who had not downloaded anything within at least the last 30 days prior to the end of the observation period. Although this assumption can be disputed, its error is nonetheless limited as it only affects inactive clients who dropped the subscription tightly near the end of the observed period. About 90\% of the inactive users ended the subscription some time before (even years before) the end-date of the time series, and are thus not affected by this cutoff. There is also the possibility of false negatives as some potential inactive users may have yet to churn by the end-date of the time series. \newline By the same considerations as above we determine that their number must be significantly small and thus we proceed annotating samples on that basis, in the absence of real customer contractual data, which has not been made available for this study. The inactive accounts were shifted in time to synchronise the entire data set obtaining  downloads versus number of days since last download event happened. We therefore could describe input data as $X= \{X_t \;\vert \; t\in \{1,2,\dots ,k\}\}$ sets (time series), where $t$ denotes the observation times. 

This way we define a binary supervised learning classification problem, where the targets (Active and Inactive) are to be inferred from the features (volume of FTDs during $n$ time before the eventual cancellation). That is, for each customer $x$, we aim at inferring $y_x$ target, where $y \in\{+1, -1\}$ is either Active (+1) or Inactive (-1). For this the hypothesis for learning can be defined at time $t$ as follows:

\begin{equation}
h_t(x) = (FTD_x, t-n)
\label{eq:1}
\end{equation}

\noindent where $FTD_x$ is the volume of downloads ($|FTD|$) for customer $x$ at time $t-n$. We argue that the behaviour of customers who are likely to cancel a subscription in the months before they effectively do so is different from that of continuing customers. This subtle difference could --however-- be learned (identified) by a trained algorithm. 

The volume of downloads relative to each customer was subsequently down-sampled, and the learning algorithms were trained on re-sampled $FTD_x$, where a re-sample ($\tilde{x}$ for simplicity to refer to $FTD\tilde{x}$) can be given by:

\begin{equation}
\tilde{x}_\iota = \sum_{k\in \Gamma_\iota} x_k
\end{equation}

\noindent where $\Gamma_\iota$ are sets composed of consecutive drawn indices. $\Gamma_\iota = 2,\dots, 30$. That is, the process is repeated both for several values of the down-sampling (from 2-day to 30 day averages) as well as for a set of prediction windows ( $t - n$ with $n$ ranging from 0 days to 365 days), considering that prediction not only needs to be accurate but also timely. The process therefore can be structured as in the pseudo-code of Alg. \ref{alg1} below:

\begin{algorithm}[H]
\label{alg1}
\KwData{$x$, lag $n$, $t, \iota$, clf \Comment{<clf: classifier, $\iota$: re-sampling index, $t$: prediction time>}}
\KwResult{clf.accuracy ($\tilde{x}, n$) }
$\iota \leftarrow 2$\;
$n \leftarrow 0$\;
\While{$\iota \leq 30$}{
 $\tilde{}{x} \leftarrow \sum_{k=\iota} x_k$\;
\While{$n \leq 365$}{
$clf_t(x) = (\tilde{x}, t-n)$\;
record accuracy ($clf_t(x)$) \Comment{<cross validation result for classification>}
$n \leftarrow n+1$
}
 $\iota \leftarrow \iota+1$\;
Exit\;
  }
\caption{Resampling-lag method}
\end{algorithm}

\noindent The accuracy of the the trained algorithms was calculated using a 5-fold cross validation. 
Only \textit{conventional} machine learning methods for binary classifications were considered\footnote{Deep learning algorithms were not considered at this stage, although their performance can be compared to evaluated ones here in a future work.}, namely Support Vector Machine (SVM), Naive Bayes and a multilayer perceptron (MLP) neural network. We additionally examined ensemble methods of Random Forest and Gradient Boosting algorithms, all of which have been extensively evaluated in the context of customer churn prediction application as in the examples provided in Section \ref{sec:1}.

It is worth mentioning that even though the data presented here are time series, we have discarded a time-series specific approach for the task. A popular non-AI method of characterising time-series is called 'Shapelets' (\cite{lines2012shapelet}) and it is widely used in pattern recognition algorithms of e.g. EEG\cite{eeg}, allowing for the identification of a time series sub-sequence as being representative of class membership, which can be elongated or moved in time and still be recognizable. As the traditional methodology (CPD) is not based on time-series, we are pursuing the next incremental step towards better understanding of the problem.  

One further reason to consider, on the choice of learning methods, is that time series analysis requires factoring in trend and seasonality components and many more features that may or may not be relevant to an effective customer churn prediction. Many complex time series classification methods are available elsewhere (e.g. \cite{geurts2001pattern,bagnall2017great}), although in many learning tasks, including the application of interest here, these may not be necessary, as non-time series --and simpler-- methods can be up to the mark. 

\section{Results and Discussion}
\label{results}
\subsection{CPD as a predictor}

\begin{figure}[h]
	\centering
		\includegraphics[width=0.8\linewidth]{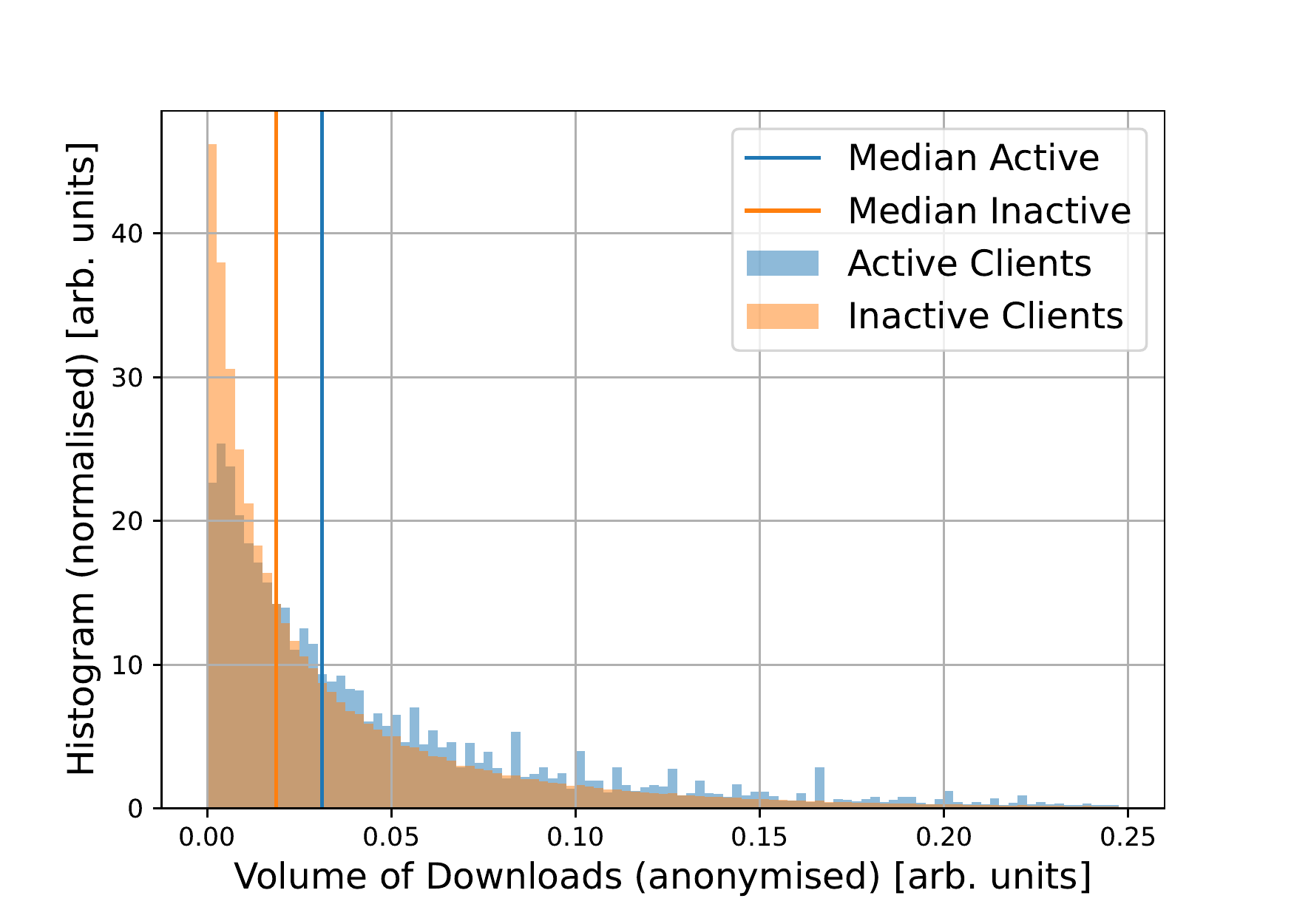}
	\caption{Statistical comparison of the population of active and inactive users by means of normalised histograms. Days of zero FTDs have not been considered. Active users tend to download more content per day, however both populations are not significantly different, from a statistical point of view. As a result, predictions of customer churn based on statistical properties will only work in extreme cases (like the orange peak near the origin) and not necessarily in general. }
	\label{FIG:CPD}
\end{figure}

CPDs, as discussed earlier is defined as the total cost of the subscription divided by the volume of downloads in a given time period. As we do not have the cost of the subscription for each client, we cannot directly evaluate CPD. However we can study the statistical properties of both populations of active and inactive samples, in order to establish if it is possible to distinguish them statistically. In figure \ref{FIG:CPD} the distributions of the volume of anonymised downloads for both populations is shown. As it is expected, both follow a fat-tail distribution, by which it seems evident that the active clients tend to download more content than the churners. Both medians are indicated as vertical lines on the figure and it can be observed that the median for active users is slightly higher than that for inactive ones. The two distributions are ---however--- very interrelated (they overlap), and there would be no statistical approach to separate them hence we are unable to reject the null hypothesis.  

We have verified that --to the first order-- institutions which download a huge amount of content are very unlikely to cancel their subscriptions, hence supporting the claim that  CPD can be used as a first order predictor. The reasons can be many but we believe it is logical for an institution which makes thousands of daily downloads of content to be of a scale that allows it to continue the subscription, and to be willing to do so as it has a validated interest from its users. The real challenge --however-- is the prediction in cases where the number of downloads are moderate.

\subsection{Ensemble learning: Random Forest}

\begin{figure}[h]
	\centering
		\includegraphics[width=0.7\linewidth]{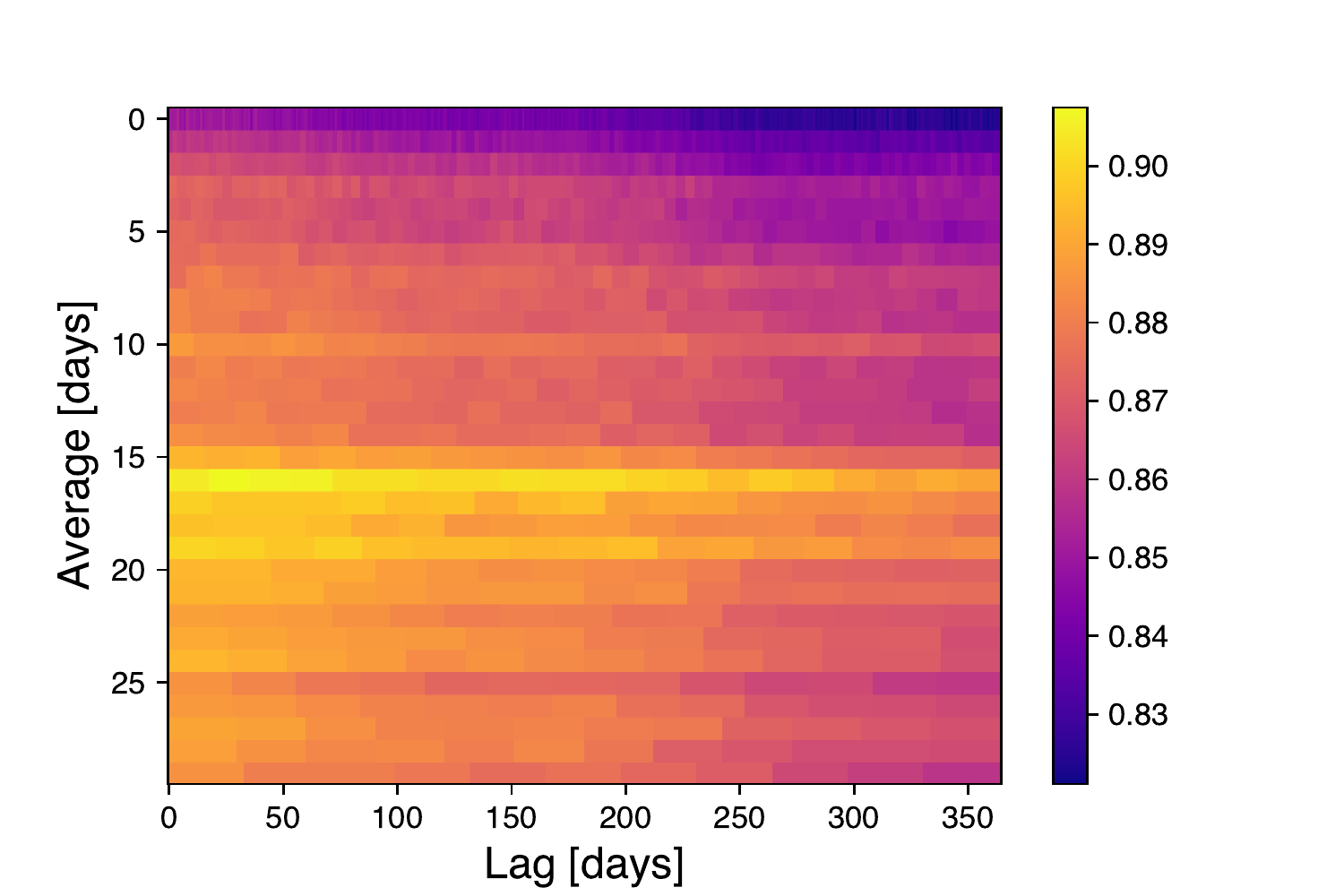}
	\caption{Map showing the accuracy using the Random Forest algorithm for the whole system as a binary classifier of customer churn. The yellows show performance (accuracy) of over 90\% . Even the worst prediction is above 80\%. The vertical axis shows the dependence of accuracy with averaging (re-sampling) the data while the horizontal axis takes days out of the data set to infer the predictive power of the setup. }
	\label{FIG:surface}
\end{figure}

The Random Forest algorithm (with 300 trees, GINI method, and the \textit{out-of-bag} (OOB) estimate for error \cite{oob}) was trained and evaluated for a wide range of scenarios, as shown in Fig. \ref{FIG:surface}. The Accuracy is shown on a colour scale where it can be observed that the way the data is averaged is a better determinant of model accuracy than the prediction window ($n$ as defined in Eq. \ref{eq:1}).
Model accuracy seems to peak at 15-20 days of average FTDs and the peak mostly persists towards even a year ahead prediction. On the contrary as non-averaged data was used, performance bottomed. It additionally worsens for higher values of time-lag, which is to be expected as the algorithm has less and less-to-date information to learn for the prediction.

If we make several cuts of figure \ref{FIG:surface} in order to study model performance with respect to the given variables we get the panels shown in figure \ref{FIG:cuts}. On the left panel we see how accuracy peaks at 17-day average of FTDs, whilst seemingly independent of the time-lag applied. We additionally observe that up until approximately a 90-day lag, model performance was nearly constant, whereas it degrades significantly for lags of 180 and 365 days. This indicates that it is not the activities of the customers during the last months that are determinant of customer churn, but rather the overall use of the resource. To examine this further we show on the right panel of fig. \ref{FIG:surface} how model performed with respect to the different time-lags ($n$) on different averaging (re-sampling) of FTDs. We can see that the peak at 17 days of averaging is consistent for all lags and the worst recorded accuracy is for models trained on the non-averaged data set. This is possibly because fine-grained resolution of FTDs seems to include noise interfering with information signal that is required by the model to accurately infer target outcomes, which seems to be embedded in a lower feature space. As is to be expected, the resolution of the averaged curves cannot be higher than the value of the averages, thus explaining the lack of high frequency noise in the averaged plots.

\begin{figure}
	\centering
		\includegraphics[width=0.49\linewidth]{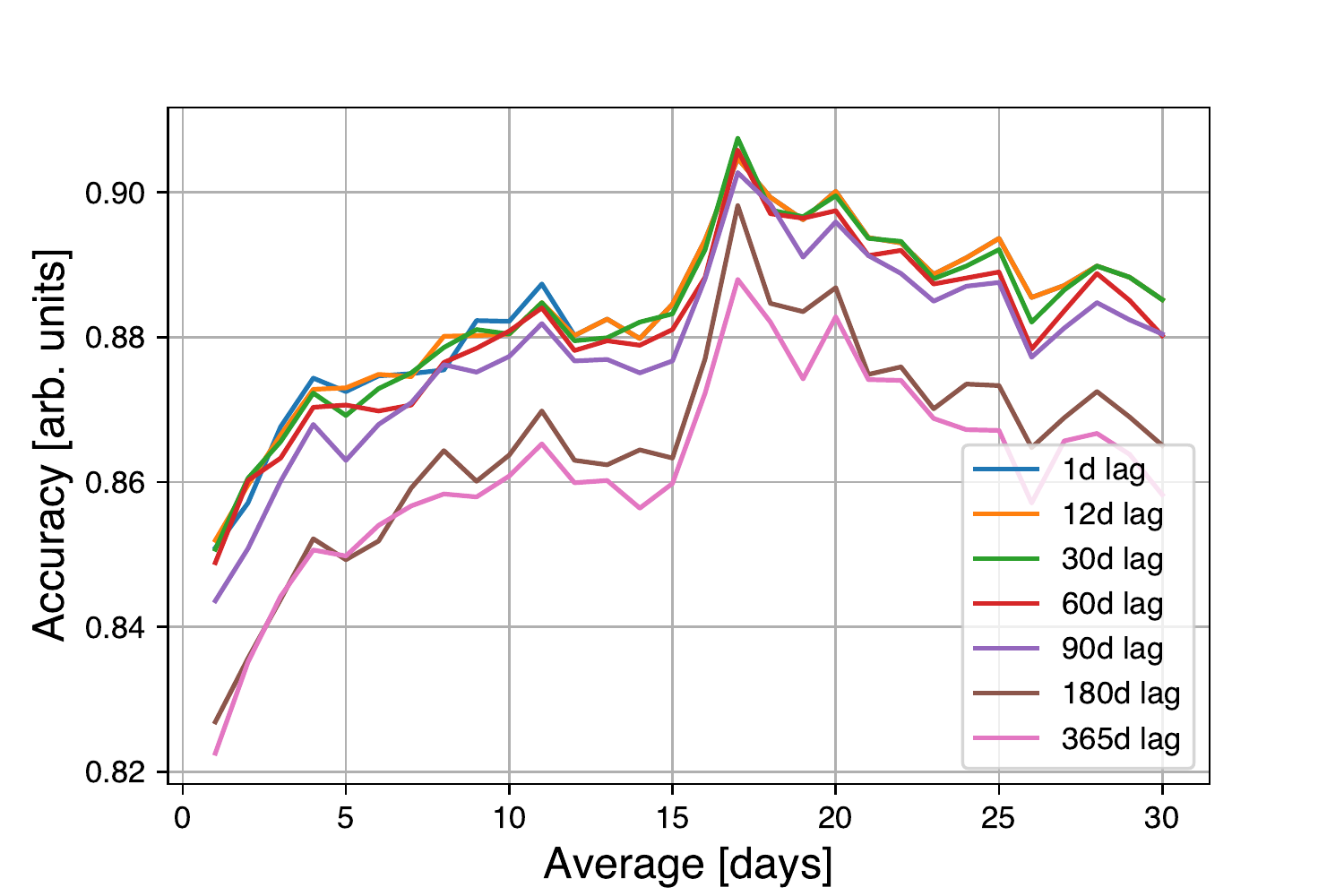}
		\includegraphics[width=0.49\linewidth]{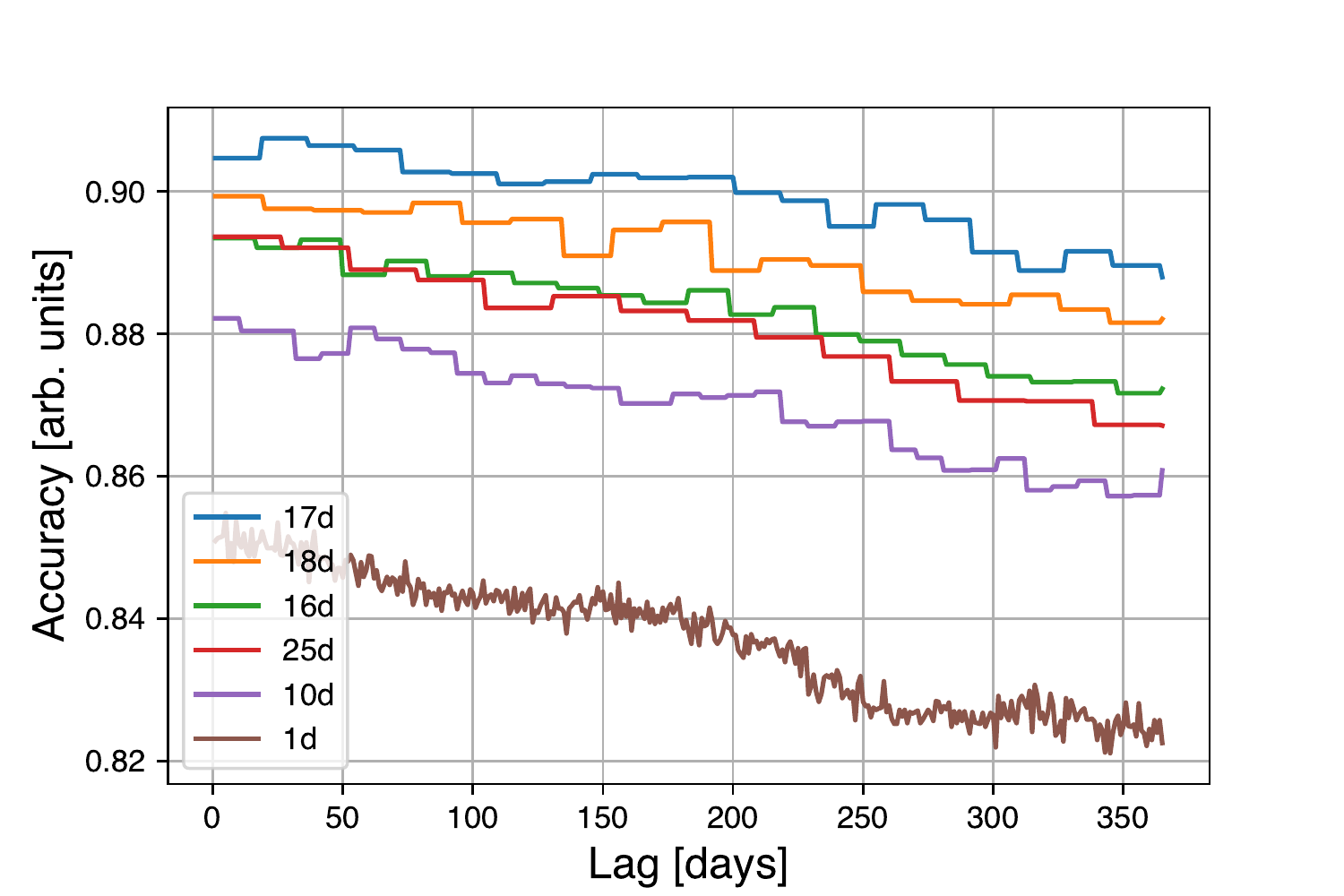}
	\caption{(left) Performance of the Random Forest relative to re-sampling (excerpt from figure \ref{FIG:surface}), showing that the optimal average is of 17 days. Also lags under 90 days have little effect on the overall accuracy of the system, giving a 90-day window for prediction. (right) Model performance relative to different lags (days for prediction) and re-samplings, showing that it consistently peaks at average of 17 days for all lags and that accuracy slowly diminishes as the lag increases.}
	\label{FIG:cuts}
\end{figure}

We additionally evaluated the rest of models in order to compare the accuracy of the proposed approach for prediction across the different classifiers. These included another ensemble learner, Gradient Boosting, a Support Vector Machine (SVM), Naive Bayes and a multi-layer perceptron (MLP) network, all for binary classification. Both ensemble algorithms, RF and Gradient Boosting, outperformed the rest of classifiers whereas all in general recorded high prediction skill, consistently so, additionally, at re-samples around 17 days of FTDs.  
\begin{figure}
	\centering
		\includegraphics[width=0.7\linewidth]{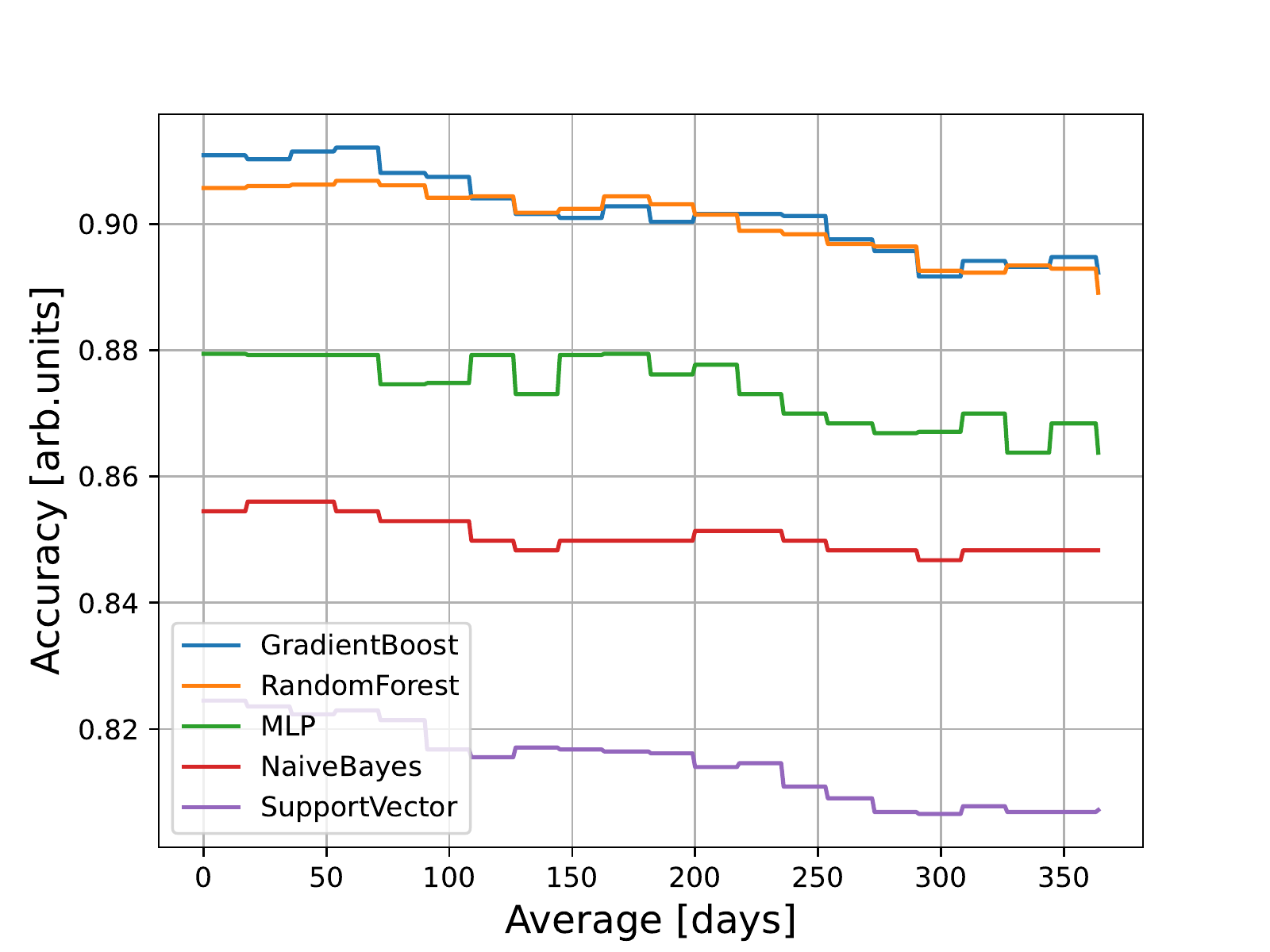}
	\caption{Comparison of the different algorithm performance for the maximum average of 17 days, with respect to lags for 365 days. The algorithms in order are: Gradient Boosting, Random Forest (also shown in previous figures), Multi-layer Perceptron (MLP), Naive Bayes and finally Support Vector Machines. The only algorithm performing comparably as well (and in some cases better) as Random Forest  is the Gradient Boosting algorithm. }
	\label{FIG:cuts}
\end{figure}

\section{Alternative Approach: Moving Averages}

An attempt to make predictions using moving averages was made. The rationale behind this is that normal averaging generates under-sampling with potential loss of information on one side, and that moving averages is able to keep relevant information and low frequency variations while keeping the number of data points (quasi) intact, on another. 

\begin{figure}
    \centering
    \includegraphics[width=0.7\linewidth]{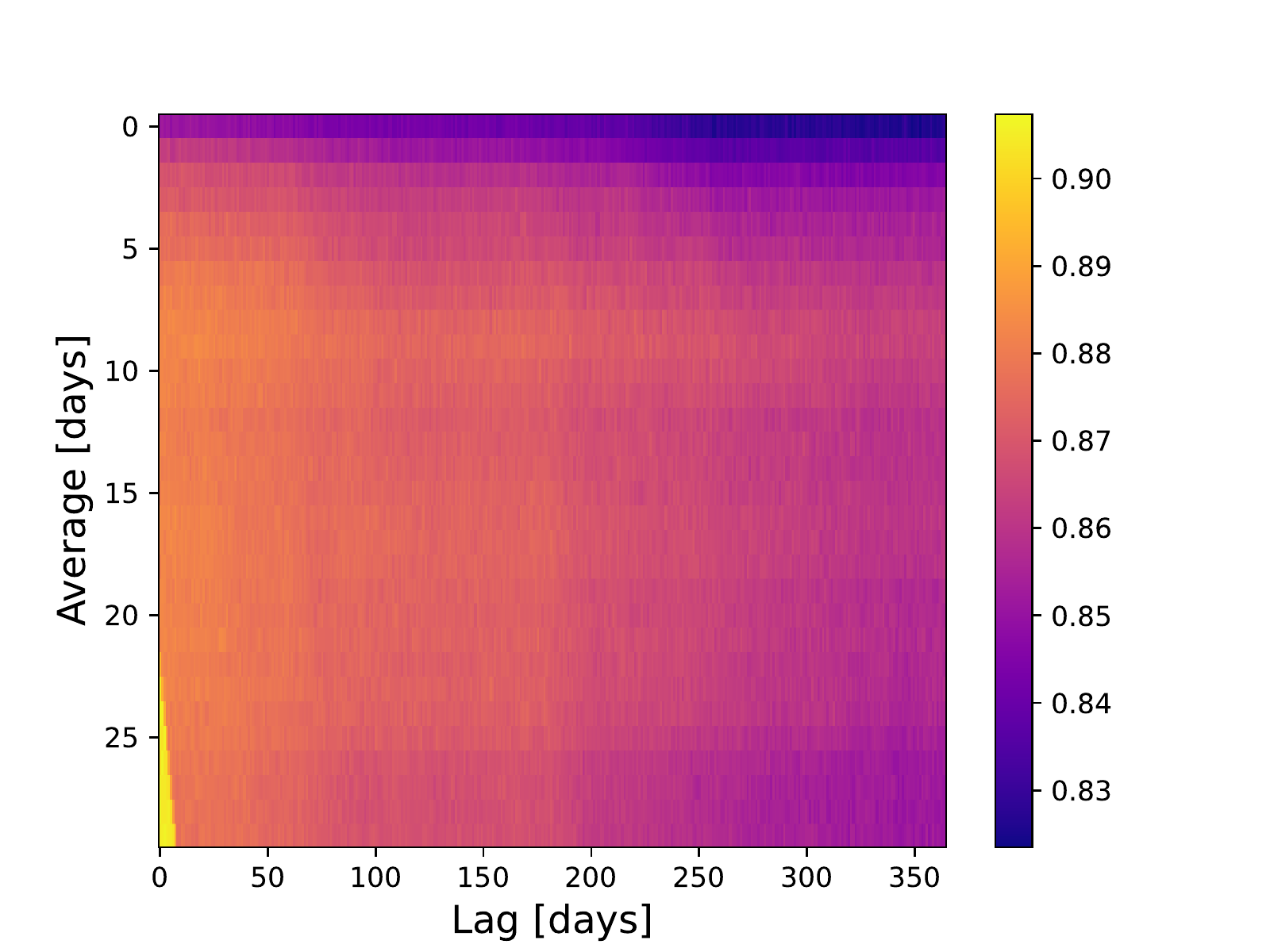}
    \caption{Map showing the accuracy using the Random Forest Algorithm as a binary Classifier as in Fig. \ref{FIG:surface}, this time instead of down-sampling the data, moving averages with a square window of length $L$ was used. The causes of relative under-performance here can be reduced to the high dimensionality of the problem and the fact that many features are very similar to one another. That is, inference is to be learnt on a singular matrix which is known to hinder model generalisability. }
    \label{fig:movAvg}
\end{figure}

Modelling results using moving averages, as shown in figure \ref{fig:movAvg}, are generally less accurate than models trained on under-sampled data. In only a handful of cases accuracy was over 90\%, although these cases give almost no prediction time-frame for a proactive action for retention.  We argue that this behaviour is due to two factors:

\begin{itemize}
    \item Even if the moving averages is able to reduce the noise, machine learning algorithms are still trained to learn and infer a very high dimensionality problem.
    \item Many of the features of the averaged data are essentially the same, this being more so the bigger the moving average window is. Training a machine learning model on a highly singular matrix, where many features are essentially duplicates of one another, leads to models with high variance which often tend to under-perform.
\end{itemize}

Consistent with the Nyquist–Shannon sampling theorem \cite{nyquist1928certain}, the information contained in sampled data can be reconstructed exactly if its maximum frequency is lower than half the Nyquist limit. Moving averages effectively decreases the maximum frequency and thus it would be possible to down-sample the data without further information loss (other than the moving averages themselves). We find this approach worthy of further exploration in a future work.

\section{Conclusions}
\label{conclusion}

This study aims to examine machine learning methods, and presents a re-sampling approach to modelling, for churn prediction in the scholarly publishing market. To do so we analyzed a data set of daily service-consumption for 10279 subscription users of a global scholarly publishing house, over a period of 6.5 years, presenting as such the first empirical analysis of customer churn in the scholarly publishing market. The paper additionally highlights the importance of this type of research in the context of B2B market: to help businesses retain their existing customers and therefore sustain their revenue, by taking proactive and timely measures to reduce the risk of profit loss. Such responsive measures rely predominantly on accurate and timely prediction of customers' likelihood to defect. 

The proposed method for prediction attempted to provide inference of customer's likelihood of defection by machine learning models trained on down-sampled data of their use of provider resources. This is motivated by two main assumptions: (1) machine learning models can be more effective on learning a lower-dimensional representation of the problem, and (2) a down-sampling approach to consumption behaviour can encompass sufficient information about its variability, without inducing significant information loss. This can be equivalent to an adaptive screening of features of high importance in a features engineering process. We examined the proposed approach considering variable time-lags for prediction by training a set of machine learning classifiers, including Support Vector Machine (SVM), Naive Bayes and a multilayer perceptron (MLP) classifiers, as well as the ensemble methods of Random Forests and Gradient Boosting. All predictive models examined, especially ensemble methods, achieved highly accurate prediction of churn, nearly a year ahead. Precisely, at 17-day re-samples of customer consumption data, the random forest algorithm achieved a 90\% prediction accuracy towards a 200-day lag (i.e. 200 days ahead), and above 85\% accuracy for a year ahead. On the contrary, when a higher-frequency (higher-dimensional representation of) consumption variability was considered, all evaluated models under-performed. This is to suggest that the modelling of churn on the basis of re-sampling customers' use of resources over subscription time is a better (simplified) approach than the typical method (learning fine-grained variability in FTDs) considering the high granularity that can often characterise individual consumption behaviour. As this allows for highly accurate and timely inference for churn from minimal possible data, further customer-oriented analysis --and related data-- can be extraneous. 

%\appendix
%\section{Data Structure}
%Appendix sections are coded under \verb+\appendix+.

\bibliographystyle{ACM-Reference-Format}

\bibliography{biblio}

\begin{acks}
HI would like to thank Miao Liu for fruitful conversations.
\end{acks}



\end{document}